\documentclass{article}
\usepackage{picins}
\usepackage{spconf,amsmath,graphicx}
\usepackage[table,xcdraw]{xcolor}
\usepackage{utfsym}

\usepackage[T1]{fontenc}
\usepackage[utf8]{inputenc}

\usepackage{enumitem}
\setlist{nosep, leftmargin=14pt}

\usepackage{mwe} 
\usepackage{multirow}
\usepackage{graphicx}

\title{A digital eye-fixation biomarker using a deep anomaly scheme to classify Parkisonian patterns}

\name{Juan Niño$^{\star}$, 
Luis Guayacán $^{\star}$, 
Santiago Gómez$^{\star}$ 
and Fabio Martínez$^{\star}$}

\address{$^\star$ Biomedical Imaging, Vision, and Learning Laboratory (BIVL$^2$ab) -\\ Universidad Industrial de Santander (UIS), Colombia}

\begin{document}

\maketitle

\begin{abstract}
Oculomotor alterations constitute a promising biomarker to detect and characterize Parkinson's disease (PD), even in prodromal stages. Currently, only global and simplified eye movement trajectories are employed to approximate the complex and hidden kinematic relationships of the oculomotor function. Recent advances on machine learning and video analysis have encouraged novel characterizations of eye movement patterns to quantify PD. These schemes enable the identification of spatiotemporal segments primarily associated with PD. However, they rely on discriminative models that require large training datasets and depend on balanced class distributions. This work introduces a novel video analysis scheme to quantify Parkinsonian eye fixation patterns with an anomaly detection framework. Contrary to classical deep discriminative schemes that learn differences among labeled classes, the proposed approach is focused on one-class learning, avoiding the necessity of a significant amount of data. The proposed approach focuses only on Parkinson's representation, considering any other class sample as an anomaly of the distribution. This approach was evaluated for an ocular fixation task, in a total of 13 control subjects and 13 patients on different stages of the disease. The proposed digital biomarker achieved an average sensitivity and specificity of 0.97 and 0.63, respectively, yielding an AUC-ROC of 0.95. A statistical test shows significant differences (p $<$ 0.05) among predicted classes, evidencing a discrimination between patients and control subjects.

\end{abstract}


\section{Introduction}
Ocular movements have emerged as a promising biomarker of Parkinson's Disease (PD) with strong diagnostic reliability \cite{iranzo2016idiopathic}. For instance, patients with alterations in rapid eye movement during sleep have shown a high risk to develop PD. More specifically, Gorges \textit{et al.} identified subtle abnormalities in ocular movement variation among PD patients, which were linked to cortical deficiencies. \cite{gorges2016association}. Also, Merel \textit{et al.} 2017 discussed various common ocular disorders that may be associated with the development of the disease, such as dry ocular surfaces, diplopia, glaucoma, colored vision and damaged contrast, visual hallucinations, among many others \cite{ekker2017ocular}. Pierpaolo \textit{et al.} 2018 also found several of these ocular abnormalities in the early stages of PD, indicating a strong association between the oculomotor function and the neurodegeneration of brain processes \cite{turcano2019early,goldberg2013}. 

Regarding ocular fixation, Gitchel \textit{et al.} reported the occurrence of ocular oscillations with frequencies around 4 to 7 Hz, similar to tremors of those with PD \cite{gitchel2012pervasive}. Besides,  Kaski  et  al. showed  that  ocular  tremor  possessed  spectral  properties  identical  to  those  of  head tremor  and  that  ocular  movement  occurred  at  180 degrees  out  of  phase  with  the  head  tremor \cite{kaski2013ocular}. In  conflict  with  this result,  Gitchel  et  al. measured  eye  and  head  movements  in  62  of  their  PD  patients  and  in  31  control subjects  and  found  that  head  movements  did  not contribute  to  the  ocular  instability  findings. Gitchel \textit{et al.} also reported that pervasive ocular  tremor is present in patients with PD. This finding has generated substantial interest, and could provide a useful biomarker for this pathology. Hence, the introduction of new technologies to characterize such emerging biomarkers are fundamental to discover new relationships associated with PD.

The use of deep learning representation methods have improved the diagnosis of PD by the modeling of diverse kinematic symptoms \cite{belic2019artificial}. Specifically, related works have reported significant discrimination between PD and control patterns applying machine learning over sensor signals on the body \cite{caramia2018imu,abdulhay2018gait}. Deep learning methods have also been used in diverse applications, such as gait analysis \cite{hu2019vision,ajay2018pervasive}. These schemes have mainly worked on the classification and recognition of gait in PD and tremor behavior, focusing on superior and inferior limbs.

\begin{figure*}[h!]
    \centering
    \includegraphics[width=0.93\textwidth]{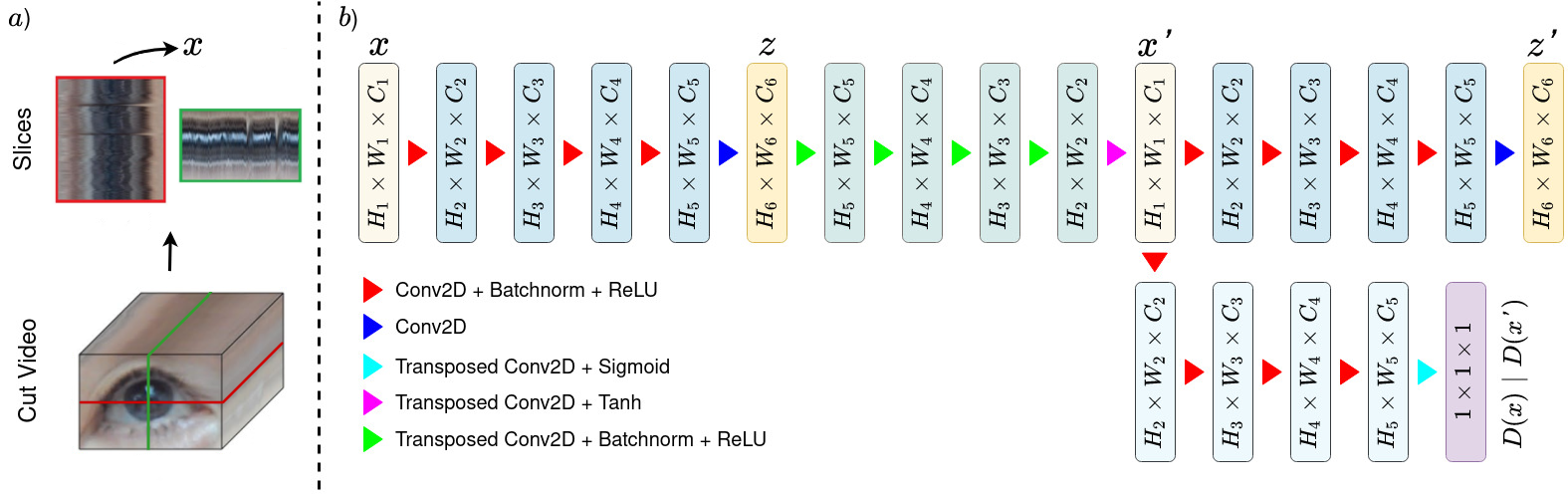}
    \caption{Overview of the proposed approach. (a) Transformation of video recordings into horizontal and vertical slices for feature extraction. (b) Anomaly detection framework based on GANomaly, where a generative autoencoder learns Parkinsonian oculomotor fixation patterns, and deviations from this learned representation are classified as outliers. The dimensions \( H \), \( W \), and \( D \) represent the height, width, and number of feature maps at different computational levels.}
    \label{fig:pipeline}
\end{figure*}

Despite the advances made in recent years, current systems for the quantitative oculomotor exam do not represent all of the ocular movement field and its intrinsic deformations. Additionally, if the equipment is expensive and difficult to mount and calibrate, it's necessarily limited to laboratory research and restrictive protocols. Other alternatives perform the oculomotor evaluation using consumption grade cameras, but ignore the subtle alterations that prove to be important to assist in the early diagnosis of PD and the tracking of the illness progression. These issues can limit the work in various scenarios with high variability like the PD, where finding oddities within these variations can be crucial. The detection of such abnormalities is an emerging field of study, where the use of generative schemes such as AnoGAN, EGBAD and GANomaly have shown positive results, with the later one achieving the best performance \cite{di2019survey,schlegl2017unsupervisedanomalydetectiongenerative,zenati2019efficientganbasedanomalydetection,akcay2019ganomaly}.

\section{Proposed approach}

This study presents two deep generative models, GANomaly and AnoGAN, to capture complex oculomotor patterns in individuals with PD and classify new cases based on their similarity to the learned representation. The proposed models facilitate the detection of significant differences in ocular fixation videos without relying on a parametric distribution to identify spatiotemporal regions associated with the disease.

\subsection{Data Collection and Processing}

In this study, we recorded video fixation motion patterns in a total of 13 PD patients (mean age of 72.3 $\pm$ 7.4) and 13 control subjects (mean age of 72.2 $\pm$ 6.1). PD patients were diagnosed in the second (5 patients), third (6 patients), and fourth (2 patients) stages of the disease by a physician using standard protocols of the Hoehn-Yahr scale. Participants were invited to observe a fixed spotlight projected onto a screen with a dark background. A conventional camera, Nikon D3200, with a spatial resolution of 1280 $\times$ 720 pixels and a temporal resolution of 60 fps was fixed in front of the subjects to capture the whole face. The eye region was manually cropped (210 $\times$ 140 pixels) to obtain the sequences of interest. A total of 130 sequences were recorded for each eye, i.e., 5 samples per person, with a duration of 5 seconds each. This study was approved by the Ethics Committee of Universidad Industrial de Santander in Bucaramanga, Colombia. Participants were recruited from the local Parkinson foundation FAMPAS (Fundación del Adulto Mayor y Parkinson Santander) and the local elderly institution Asilo San Rafael. Written informed consent was obtained from every participant. 

We extracted a vertical and horizontal slice along the center of each video (as depicted in Figure \ref{fig:pipeline}-a), producing 260 images for control subjects and 260 for PD patients. To improve dataset variability for model training, we applied data augmentation techniques, including translation, rotation, mirroring, rescaling, and noise generation. This process expanded the dataset to 800 PD training images.

\subsection{Anomaly Classification Scheme}

The proposed method learns Parkinsonian oculomotor fixation patterns to capture inter-class variability and define a disease signature, as shown in Figure \ref{fig:pipeline}. To achieve this, we employ an anomaly detection framework based on GANomaly \cite{akcay2019ganomaly}, a generative model with three subnetworks: a generative autoencoder, an encoder, and a discriminator. The autoencoder reconstructs input data by mapping an image \( X \) to a latent representation \( z \) and generating a reconstruction \( X' \), where \( z = G(X) \) and \( X' = G(z) \). The encoder \( E \) further compresses \( X' \) into a latent representation, given by \( z' = E(X') \). Finally, the discriminator \( D \) classifies the original input \( X \) and its reconstruction \( X' \) as real or fake, refining the model's ability to distinguish Parkinsonian from non-Parkinsonian patterns.

This framework, trained on Parkinsonian oculomotor patterns extracted from horizontal and vertical video slices, models a distribution \( x_P \), where control samples yield higher anomaly scores (\( x_n \gg x_P \)), thereby identifying them as outliers. Since the model is trained exclusively on Parkinsonian data, it cannot reconstruct features specific to control samples. As a result, the reconstructed output \( X' \) aligns with the Parkinsonian representation, omitting distinguishing characteristics of control data. This discrepancy causes the encoder \( E \) to generate a latent representation \( z' \) that deviates from \( z \). When this deviation exceeds a predefined threshold, the model classifies \( X \) as an outlier. To quantify this effect, we define an objective function composed of three loss components: the encoder loss \( \mathcal{L}_\text{enc} = || z - z' ||_2^2 \), which measures latent space discrepancy; the contextual loss \( \mathcal{L}_\text{ctx} = || X - X' ||_1 \), ensuring pixel-wise reconstruction fidelity; and the adversarial loss \( \mathcal{L}_\text{adv} = || f(X) - f(X') ||_2^2 \), which enhances the discriminator's ability to detect generated samples. The final anomaly score for an input image \( X \) is computed as: 

\begin{equation}
A(X) = || G(X') - E(G(X')) ||_2^2,
\end{equation}

where a higher score indicates greater deviation from the learned Parkinsonian representation.

This anomaly detection approach enables the development of digital biomarkers that objectively distinguish Parkinsonian oculomotor signatures from those of control subjects. By leveraging a data-driven strategy, this method provides a robust framework for PD classification, supporting automated diagnosis and clinical decision-making.

\subsection{Experimental setup}
We performed 4-fold cross-validation, randomly assigning 10 patients for training and 3 for validation in each iteration. The training process was conducted using image slices from PD patients for 60 epochs, using the Adam optimizer with a fixed learning rate and a batch size of 1 \cite{kingma2014adam}. Loss contributions were empirically tuned, assigning a weight of 1 to both the encoder and adversarial networks and 50 to the contextual network.

For validation, we used 100 images from PD patients and 207 from control subjects. An anomaly score was computed for each test image. AUC-ROC, precision, and recall were used as performance metrics to compare GANomaly and AnoGAN. Additionally, statistical significance was assessed using ANOVA. The threshold for classifying an image as anomalous was empirically determined through cross-validation.

\section{Results}
The proposed anomaly detection model, based on the GANo-maly network, demonstrated strong classification performance in distinguishing PD patients from control subjects. The model achieved a mean F1-score of 0.76, which reflects a balanced trade-off between precision and recall. Specifically, the precision (the proportion of true positive predictions among all positive predictions) was 0.63, while the recall (the proportion of correctly identified positive cases among all actual positives) reached 0.97, highlighting the model's high sensitivity in identifying Parkinsonian patterns.

\begin{figure}[h!]
    \centering
    \includegraphics[width=0.48\textwidth]{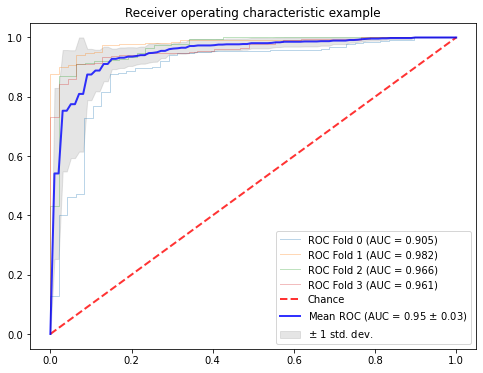}
    \caption{
    ROC curves for the different cross-validation folds of the GANomaly approach. The solid lines correspond to individual folds, while the bold blue line represents the mean ROC curve. The shaded region denotes the variability across folds, and the dashed red line indicates the performance of a random classifier.}
    \label{fig:roc}
\end{figure}

The Receiver Operating Characteristic (ROC) curve analysis further confirms the model's effectiveness. The mean Area Under the Curve (AUC-ROC) was 0.95 $\pm$ 0.03, indicating a high discriminative capacity across different validation folds. Figure \ref{fig:roc} shows the ROC curves obtained for each fold, with AUC values consistently above 0.90, reflecting robust classification performance. The red dashed line represents the chance level (AUC = 0.50), while the blue curve corresponds to the mean ROC across all folds, with a shaded region denoting $\pm$1 standard deviation.

\begin{figure}[h!]
    \centering
    \includegraphics[width=0.48\textwidth]{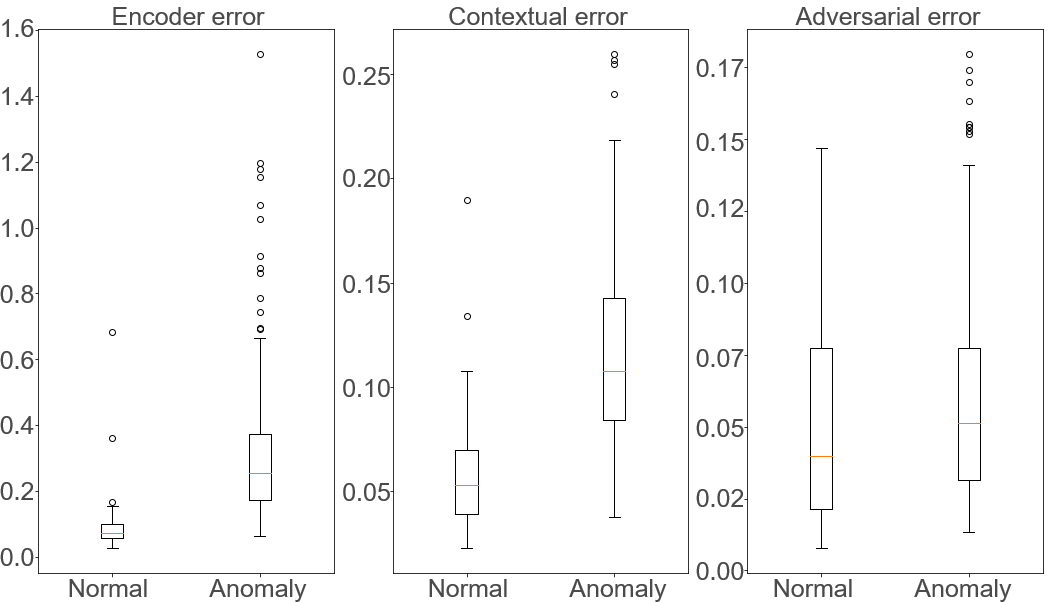}
    \caption{Boxplots comparing normal and anomalous populations across the Encoder, Contextual, and Adversarial subnetworks of the GANomaly model. The significant differences, confirmed by one-way ANOVA, highlight the model's effectiveness in distinguishing between control and PD populations.}
    \label{fig:boxplot}
\end{figure}

To enhance classification performance, we empirically determined an anomaly score threshold of 0.056, ensuring a balance between precision and recall. According to this threshold, subjects with scores below this value are classified as PD patients, while those above it are considered outliers (control subjects). This threshold selection maximized the classification performance, ensuring that the model effectively captured the underlying characteristics of Parkinsonian eye fixation patterns. Then, we generated the confusion matrix shown in Figure \ref{fig:boxplot}. Out of the three PD patients used in the validation process, two were correctly classified, while one was misclassified. Similarly, among the 13 control subjects, 12 were correctly identified, and one was incorrectly classified as a PD patient. 

Additionally, we created a boxplot to visualize the differences between both populations across each of the subnetworks, followed by a statistical analysis. In our GANomaly experiment, we computed the mean and variance for both the normal and abnormal populations in each subnetwork. To assess statistical significance, we conducted a one-way ANOVA test, with the null hypothesis stating that there is no significant difference between the two populations. If the resulting p-value was below 0.05, the null hypothesis was rejected, indicating that the model effectively distinguished between both groups.

Furthermore, we conducted a statistical comparison between the AnoGAN and GANomaly experiments, where, in the case of GANomaly, the reference and outlier classes were inverted. Since AnoGAN does not include subnetworks, the one-way ANOVA test was performed only once for this model.

For AnoGAN, the one-way ANOVA statistic was 0.02, with a p-value $>$ 0.65, meaning the null hypothesis could not be rejected. This indicates that the model failed to identify significant differences between control and PD patients. The variance difference for this network was 0.97.

In contrast, the GANomaly approach demonstrated a stronger capacity to distinguish between groups. The encoder subnetwork yielded an ANOVA statistic of 46.81 and a p-value $<$ 0.05, confirming significant differences, with a variance difference of 0.10. The contextual subnetwork achieved a statistic of 80.84, also with a p-value $<$ 0.05, and a variance difference of 0.26. The adversarial error subnetwork obtained a statistic of 1.39, rejecting the null hypothesis as well, with a variance difference of 0.79.
When reversing the reference class in GANomaly, similar trends were observed. The encoder subnetwork produced a statistic of 77.26, with a p-value $<$ 0.05, and a variance difference of 0.13. The contextual subnetwork showed the highest statistical significance, with a statistic of 248.74, a p-value $<$ 0.05, and a variance difference of 0.13. Finally, the adversarial error subnetwork yielded a statistic of 8.49, also rejecting the null hypothesis, with a variance difference of 0.18.

These results confirm that GANomaly consistently detected significant differences between the two populations, whereas AnoGAN did not. Additionally, the first GANomaly approach outperformed the second one, as reflected in its higher AUC-ROC score. This suggests that the original reference-outlier assignment (PD as the reference class and controls as outliers) provided better discrimination between the two groups.

\begin{figure}[h!]
    \centering
    \includegraphics[width=0.48\textwidth]{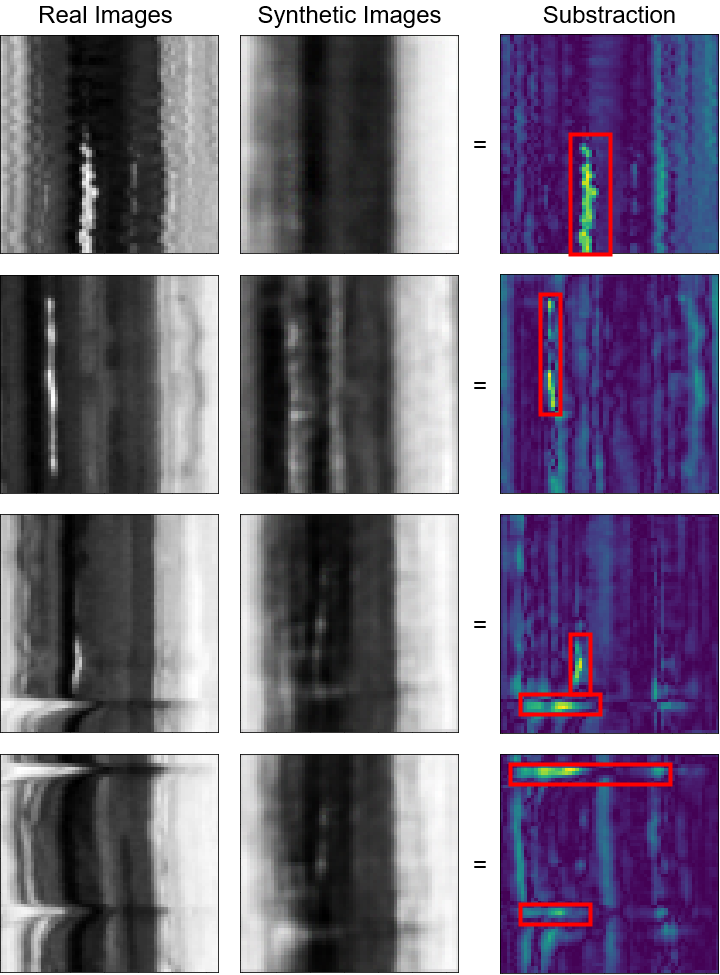}
    \caption{Subtraction of some images before and after entering the network (real and synthetic images respectively). The differences are shown in the domain of frequencies, with the relevant alterations in frequency represented by high contrast.}
    \label{fig:substraction}
\end{figure}

To further analyze the transformations introduced by the model, we performed a subtraction between real and synthetic images before and after being processed by the network. As shown in Figure \ref{fig:substraction}, the differences are represented in the frequency domain, where high-contrast regions highlight the most relevant alterations. This visualization provides insight into how the network modifies and reconstructs input data, revealing the specific frequency components that contribute to distinguishing between real and synthetic samples.

\section{Discussion and concluding remarks}
This work introduced a deep generative approach for identifying differences in ocular fixation videos to aid in the classification and diagnosis of PD. The GANomaly architecture achieved a mean ROC AUC of 0.95, significantly outperforming the AnoGAN implementation. 
With a clear rejection of the null hypothesis, GANomaly successfully distinguished between the two populations without assuming a parametric distribution over the data. This enabled the identification of spatiotemporal segments associated with PD.

During experimentation, we observed that using the PD population as the reference class in the learning process yielded better test performance than using the control group as the reference. As discussed in the previous section, we inverted the reference class in the second GANomaly experiment, which resulted in higher statistical differences but slightly lower test performance.

This biomarker identification experiment provides further support for the ocular tremor vs. head tremor hypothesis previously mentioned, as we were able to detect biomarkers using only oculomotor fixation data. No head movement information was recorded, as the analysis focused exclusively on the ocular region. 
Additionally, since our results demonstrated that this architecture benefits from larger training datasets, we expect that acquiring more data could further improve performance and precision.

\section*{Ethics statement}
The local ethics boards of all participating centers granted approval for the retrospective assessment of imaging data. To ensure complete anonymity, all patient information was eliminated from the volumetric nifty files. Due to the retrospective nature of the study and the de-identification of patient data, the requirement for written informed consent was exempted by the ethics boards.

\bibliographystyle{IEEEbib}
\bibliography{refs}







\end{document}